\documentclass[a4paper, 10pt, conference]{ieeeconf}      % Use this line for a4 paper

\IEEEoverridecommandlockouts                              % This command is only needed if 
\usepackage{graphics} % for pdf, bitmapped graphics files
\usepackage{epsfig} % for postscript graphics files
\usepackage{mathptmx} % assumes new font selection scheme installed
\usepackage{times} % assumes new font selection scheme installed
\usepackage{amsmath} % assumes amsmath package installed
\usepackage{amssymb}  % assumes amsmath package installed
\usepackage{pifont} % circled digits in the figure caption
\usepackage{booktabs} % experiment count table

\title{\LARGE \bf
Online Adaptation to Latent Physical Properties for Learning from Demonstration
}

\author{Kotaro Kajitsuka$^{1,2}$, Masaki Murooka$^{2}$, and Fumio Kanehiro$^{1,2}$% <-this % stops a space
\thanks{$^{1}$University of Tsukuba, Tsukuba, Japan.}%
\thanks{$^{2}$CNRS-AIST JRL (Joint Robotics Laboratory), IRL3218,
        National Institute of Advanced Industrial Science and Technology (AIST),
        Tsukuba, Japan. {\tt\small tactile.kajitsuka@aist.go.jp}}%
\thanks{This work was partially supported by JST CREST, Japan, under Grant Number JPMJCR2553, and JST Moonshot R\&D Program, Japan, under Grant Number JPMJMS263D.}% <-this % stops a space
}

\begin{document}

\maketitle
\thispagestyle{empty}
\pagestyle{empty}

%%%%%%%%%%%%%%%%%%%%%%%%%%%%%%%%%%%%%%%%%%%%%%%%%%%%%%%%%%%%%%%%%%%%%%%%%%%%%%%%
\begin{abstract}

Manipulating visually identical objects with different physical properties requires adapting actions based on information acquired through interaction.
We address the problem of learning from demonstrations how to adapt manipulation actions, such as regrasping and changing contact locations, based on physical-property information acquired during interaction.
To this end, we propose a learning-from-demonstration method that learns latent representations of physical properties needed to predict object responses and conditions a policy on online estimates of these representations.
A dynamics model and object-specific latent representations are jointly learned from demonstrations that include behaviors adapting to the objects' physical properties, and the learned representations are then used to condition an imitation policy.
During execution, the dynamics model and policy weights remain fixed, while the latent representation is updated based on the discrepancy between predicted and actual observations, allowing interaction with the object to inform action generation.
Real-world and simulation experiments with objects whose centers of mass differ suggest that online adaptation through latent representations of physical properties can improve task success rates.
However, in simulation, success rates on center-of-mass locations absent from the demonstrations remained around 30\%, indicating limited adaptation to unseen physical properties.
Project page: https://isri-aist.github.io/latent-dynamics-project-page/
\end{abstract}

%%%%%%%%%%%%%%%%%%%%%%%%%%%%%%%%%%%%%%%%%%%%%%%%%%%%%%%%%%%%%%%%%%%%%%%%%%%%%%%%

\section{Introduction}

Learning from demonstration (LfD) has been applied to a wide range of robotic manipulation tasks, including manipulation of deformable objects and long-horizon tasks.
These methods have shown strong performance in settings where visual observations provide sufficient information for action selection~\cite{Chi:DiffusionPolicy:IJRR2025}, while adding force measurements supports delicate tasks such as wiping~\cite{Hou:AdaptiveCompliancePolicy:ICRA2025,Choi:UMI-FT:ICRA2026}.
Humans infer physical properties such as mass and friction through interaction and use this information to guide subsequent actions, for example, adjusting their grasp and posture when lifting a heavy object.

Regrasping or changing contact locations based on inferred properties requires retaining information from earlier contacts and using it to select subsequent actions.
Policies conditioned on a short observation window may struggle when action selection requires information outside that window~\cite{Zhou:MTIL:RAL2025}.
In this work, we focus specifically on physical properties inferred from earlier interactions.

Beyond LfD, online system identification has been extensively studied through analytical approaches, dynamics-based property estimation~\cite{Murooka:FeasibilityEvaluation:ICRA2017}, and adaptation using simulation-trained dynamics models~\cite{Murooka:EXI-Net:PMLR2021}.
Connecting this body of work with LfD could substantially broaden the range of manipulation problems that can be learned from demonstrations.

In this work, we collect demonstrations of property-dependent manipulation using visually identical objects with different centers of mass (CoM) or centers of friction (CoF).
The demonstrations include corrective actions in response to unexpected object motion, such as regrasping or changing contact locations.
Within this setting, we propose and evaluate an imitation learning method that adapts online to differences in object physical properties.

\begin{figure}[t]
  \centering
  \includegraphics[width=\columnwidth]{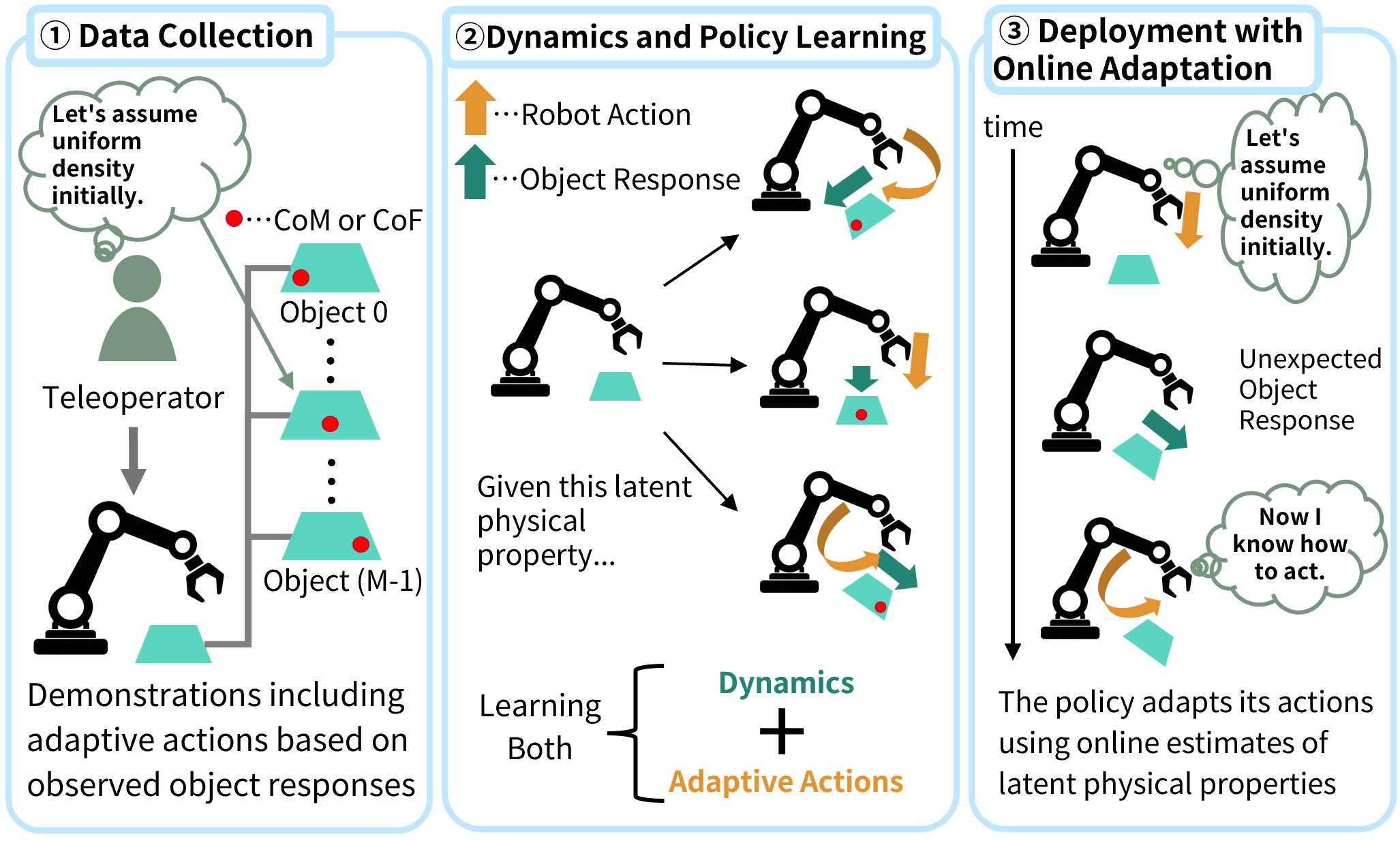}
  \caption{Overview of the proposed approach.
  \ding{172} Teleoperated demonstrations include adaptive actions based on observed object responses, such as regrasping or changing contact locations.
  \ding{173} Using the same demonstrations, we first jointly learn a dynamics model and object-specific physical representations, then train an imitation policy conditioned on the learned representations. This learns both dynamics and adaptive actions that depend on physical properties.
  \ding{174} Starting from a reference physical representation, the robot updates it from prediction errors and conditions subsequent actions on the estimate. The figure illustrates uniform density as the initial assumption.}
  \label{fig:concept}
\end{figure}

Fig.~\ref{fig:concept} provides an overview of the proposed approach.

To adapt to these latent physical properties, we introduce an implicit dynamics parameter $\mathbf d_i$ as a latent representation of object $i$.
We first learn the dynamics model and $\mathbf d_i$ from demonstrations, then train a policy conditioned on the learned representation.
During execution, we update the estimate $\hat{\mathbf d}$ online using discrepancies between model predictions and actual observations, and generate actions conditioned on the inferred properties.

The method has two advantages.
First, it initially learns a dynamics model and implicit physical representations solely from demonstrations, without prior measurements of the objects' physical properties.
Second, its continuous representation, rather than a discrete object category, is intended to support adaptation to unseen objects.

Our contributions are as follows:
\begin{itemize}
  \item We show that interaction data in manipulation demonstrations can be used to learn implicit representations of differences in object physical properties.
  \item We demonstrate that combining LfD with dynamics-based property estimation improves performance on tasks involving visually identical objects with different physical properties, including deformable objects.
\end{itemize}

%%%%%%%%%%%%%%%%%%%%%%%%%%%%%%%%%%%%%%%%%%%%%%%%%%%%%%%%%%%%%%%%%%%%%%%%%%%%%%%%

\section{Related Work}
\subsection{Physical Property Estimation for Manipulation}
Physical properties such as mass and friction affect object responses and therefore matter for prediction and control.
Analytical models support approaches such as hybrid model predictive control for planar manipulation~\cite{Hogan:HybridMPC:IJRR2020}, but require specifying physical parameters and suitable assumptions about contact and friction.
The cost of characterizing each object in advance motivates estimating its properties through interaction.

For predicting object responses, latent representations can capture relevant physical differences without explicitly estimating physical parameters such as mass or friction coefficients.
Longhini et al.~\cite{Longhini:EDO-Net:ICRA2023} learn latent representations of elasticity from pulling interactions to condition graph-based predictions of cloth dynamics.
Evans et al.~\cite{Evans:IIDA:ICRA2022} infer latent context from state transitions, including real-robot sliding interactions with different objects and surfaces, to adapt dynamics predictions.
Simulation-trained models also enable adaptation to real objects: EXI-Net~\cite{Murooka:EXI-Net:PMLR2021} updates explicit and implicit dynamics parameters from prediction errors during planar pushing, while Poke and Strike~\cite{Aoyama:PokeAndStrike:CoRL2025} learns an exploration policy and a property estimator, then uses the estimated properties to condition task execution.

Physical representations have also been incorporated into LfD.
Aoyama et al.~\cite{Aoyama:FewShotHaptic:ICRA2024} pretrain haptic representations from simulated force sequences and use a few real demonstrations to learn wiping motions adapted to sponge stiffness and surface friction.
Saito et al.~\cite{Saito:SelectAndUseTools:RAL2021} infer ingredient characteristics from multimodal observations during stirring with a wooden spatula, then use the inferred representation to select and manipulate tools.
Both methods obtain these representations through a dedicated exploration phase before task execution.
Our demonstrations instead integrate task execution with responses to unexpected object behavior, including regrasping and changes in contact location.
We update the physical representation throughout execution to support these adaptive behaviors without explicitly separating exploration from the task.

\subsection{History-Dependent Imitation Learning}
Adaptive manipulation may depend on information that is no longer directly observable: an off-center mass detected during lifting, for example, can inform a later grasp after the object has been put down.
A policy conditioned only on a recent observation window loses access to such information once it falls outside the window.
Simply extending the window does not guarantee better performance.
Sobol Mark et al.~\cite{SobolMark:BPP:RSS2026} show that limited coverage of observation histories can lead policies to rely on spurious correlations that fail under rollout distribution shift.
Their BPP method uses a vision-language model to identify keyframes corresponding to task-relevant events, reducing these correlations and improving task performance.

Recurrent approaches, including the spatial-attention model of Ichiwara et al.~\cite{Ichiwara:SpatialAttentionPointNetwork:arXiv2021} and Mamba-based MTIL~\cite{Zhou:MTIL:RAL2025}, instead encode observation history in an internal state used for action generation.
Our approach focuses this retained information on physical properties relevant to predicting object responses, using a latent representation updated from dynamics prediction errors during execution.

\subsection{World Models for Policy Learning and Adaptation}
World models predict the consequences of actions and can inform policy learning and execution.
An et al.~\cite{An:FeedbackWorldModel:arXiv2026} propose a Feedback World Model that predicts future visual features and corrects its predictions online using observation feedback to guide Diffusion Policy denoising.
For nonprehensile manipulation, Lyu et al.~\cite{Lyu:DyWA:ICCV2025} introduce DyWA, which distills a simulation-trained reinforcement learning teacher into a student that jointly predicts actions and subsequent task states.
Its history-based adaptation representation is supervised by teacher features encoding object geometry and physical parameters.

Yuan et al.~\cite{Yuan:AdaWorldPolicy:arXiv2026} integrate a pretrained Cosmos world model~\cite{Ali:Cosmos:arXiv2025}, an action model, and a force predictor in AdaWorldPolicy.
Prediction errors drive online updates to low-rank network parameters, enabling adaptation to both visual and physical domain shifts.
Our focus is on learning object-specific physical representations by minimizing prediction errors on demonstration data, without physical-parameter supervision.
Our method instead updates only the latent physical representation online, keeping the dynamics model and policy weights fixed.

\section{Proposed Method}
\label{sec:proposed_method}

\begin{figure*}[t]
  \centering
  \includegraphics[width=\textwidth]{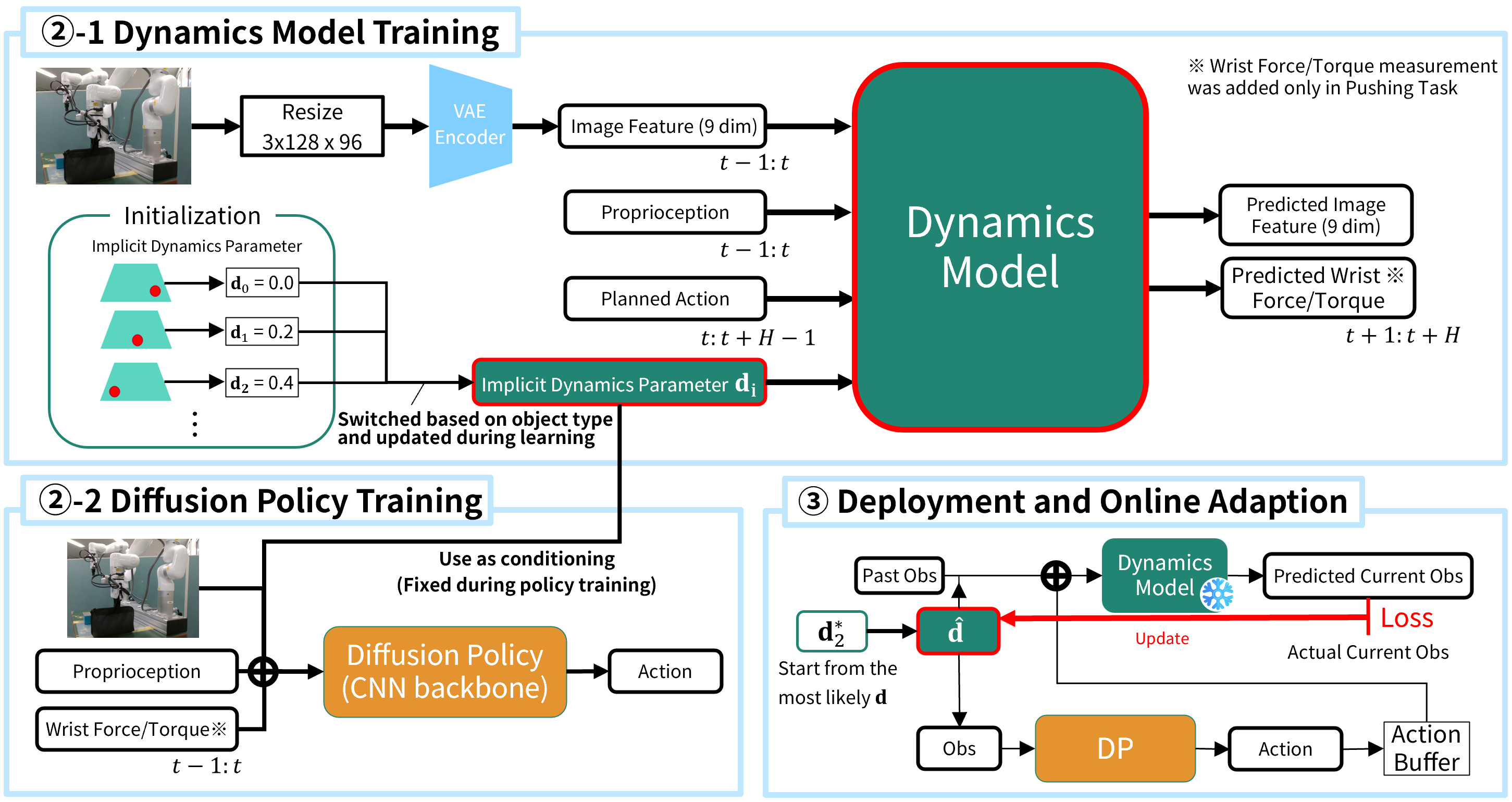}
  \caption{Training and deployment pipeline of the proposed method. $t$ denotes the discrete time-step index, and $H$ is the prediction horizon in steps.}
  \label{fig:proposed_pipeline}
\end{figure*}

\subsection{Overview}
Fig.~\ref{fig:proposed_pipeline} summarizes the proposed method.
We first collect demonstrations of manipulation adapted to the differing physical properties of visually identical objects (step~\ding{172} in Fig.~\ref{fig:concept}).
From these demonstrations, we jointly learn a dynamics model that predicts future observations and an implicit dynamics parameter representing the latent physical properties of each object (step~\mbox{\ding{173}-1} in Fig.~\ref{fig:proposed_pipeline}).
In step~\mbox{\ding{173}-2}, we annotate the demonstrations with the learned physical representations and train a Diffusion Policy conditioned on observations and these representations.
During deployment (step~\ding{174}), we keep the model and policy weights fixed, update the implicit dynamics parameter using discrepancies between predicted and actual observations, and generate actions conditioned on the current estimate.

\subsection{Data Collection, Training, and Online Adaptation}
\subsubsection*{\textbf{\ding{172} Demonstration Collection}\upshape}
\label{sec:demo_collection}
We construct a dataset $\mathcal D$ by collecting demonstrations of manipulation adapted to the differing physical properties of $M$ types of visually identical objects (step~\ding{172} in Fig.~\ref{fig:concept}).
The operator commands adaptive actions based on information acquired through interaction during manipulation.
For each demonstration, we record observations and robot action commands and assign an object-type label $i$ ($i=0,\ldots,M-1$).
We use the same dataset to train the dynamics model and the policy.

\subsubsection*{\textbf{\mbox{\ding{173}-1} Learning Dynamics and Physical Representations}\upshape}
We first train a variational autoencoder (VAE)~\cite{kingma2022autoencodingvariationalbayes} from scratch on the demonstration images (step~\mbox{\ding{173}-1} in Fig.~\ref{fig:proposed_pipeline}).
After training, we freeze the encoder and denote the low-dimensional features extracted from the image at time $t$ by $\mathbf z_t$.

Let $\mathbf d_i$ denote the implicit dynamics parameter for object type $i$.
It is a learnable embedding vector shared across all timesteps and demonstrations of that object type.
We initialize each component of $\mathbf d_i$ to $0.2i$.

Let $\mathbf p_t$ denote the robot's proprioceptive information at time $t$.
For the pushing task, the observations also include forces and torques $\mathbf w_t$ measured by a six-axis wrist force/torque sensor.

Let $\mathbf a_t$ be the action command issued to the robot at time $t$ and $H$ the prediction horizon.
The dynamics model predicts future observations from the image features and proprioceptive information at the previous and current timesteps, the action sequence over the next $H$ steps, and the implicit dynamics parameter.
The prediction target is $\mathbf y_t=\mathbf z_t$, or $\mathbf y_t=(\mathbf z_t,\mathbf w_t)$ when forces and torques are included for the pushing task.
We implement the dynamics model as a multilayer perceptron (MLP):

\begin{equation}
\label{eq:dynamics_prediction}
\hat{\mathbf y}_{t+1:t+H}
= f_\theta\left(\mathbf z_{t-1:t},\mathbf p_{t-1:t},
\mathbf a_{t:t+H-1},\mathbf d_i\right).
\end{equation}

Here, $\hat{\mathbf y}$ denotes the predictions and $\theta$ the model parameters.
Training minimizes the mean squared error between the predictions and the corresponding actual observations.
Omitting the prediction-horizon indices $t+1:t+H$, the prediction loss is

\begin{equation}
\label{eq:dynamics_loss}
\mathcal L_{\mathrm{dyn}}(\theta,\mathbf d_i)
=\operatorname{MSE}(\hat{\mathbf z},\mathbf z)
+\lambda_w\operatorname{MSE}(\hat{\mathbf w},\mathbf w).
\end{equation}

Here, $\operatorname{MSE}$ denotes the mean squared error over all timesteps and components in the prediction horizon, and $\lambda_w$ weights the force/torque prediction error.
We omit the second term for tasks without force/torque prediction.
For each demonstration, we select $\mathbf d_i$ using its object label $i$ and jointly optimize $\theta$ and $\{\mathbf d_i\}$ to minimize the prediction loss:

\begin{equation}
\label{eq:dynamics_training}
\left(\theta^*,\{\mathbf d_i^*\}\right)
=\arg\min_{\theta,\{\mathbf d_i\}}
\mathbb E_{\mathcal D}
\left[\mathcal L_{\mathrm{dyn}}(\theta,\mathbf d_i)\right].
\end{equation}

Thus, the learned physical representations capture object differences relevant to observation prediction, without requiring explicit physical parameter values.

\subsubsection*{\textbf{\mbox{\ding{173}-2} Policy Learning with Physical Representations}\upshape}
We annotate the demonstrations of object $i$ with its learned physical representation $\mathbf d_i^*$ from~\eqref{eq:dynamics_training}, yielding a dataset $\mathcal D_{\mathrm d}$ (step~\mbox{\ding{173}-2} in Fig.~\ref{fig:proposed_pipeline}).
Let $\mathbf o$ denote the observation history supplied to the policy and $\mathbf a$ the demonstrated action sequence.
We train a Diffusion Policy~\cite{Chi:DiffusionPolicy:IJRR2025} $\pi(\mathbf a\mid\mathbf o,\mathbf d_i^*)$ conditioned on the observation history and $\mathbf d_i^*$.
During this stage, $\mathbf d_i^*$ remains fixed and only the policy is trained.

The imitation learning objective can be expressed in terms of the conditional action distribution as
\begin{equation}
\label{eq:policy_training}
\pi^*
=\arg\min_{\pi}
\mathbb E_{(\mathbf o,\mathbf a,\mathbf d_i^*)\sim\mathcal D_{\mathrm d}}
\left[-\log\pi(\mathbf a\mid\mathbf o,\mathbf d_i^*)\right].
\end{equation}

\subsubsection*{\textbf{\ding{174} Deployment and Online Adaptation}\upshape}
Finally, we deploy the learned policy to control the robot and update the physical representation online using observations acquired through interaction (step~\ding{174} in Fig.~\ref{fig:proposed_pipeline}).
Let $\hat{\mathbf d}$ denote the estimated physical representation of the object being manipulated.
We initialize it with a learned $\mathbf d_i^*$ corresponding to a plausible assumption about the object's physical properties based on its appearance, such as uniform density or a center of friction at the center of the contact area.
Given the observation history and the current estimate, the learned policy $\pi^*(\mathbf a\mid\mathbf o,\hat{\mathbf d})$ generates an action sequence for the robot to execute.

We store executed action commands in a buffer and input them to the dynamics model $f_{\theta^*}$ together with past image features, proprioceptive information, and $\hat{\mathbf d}$.
The model predicts observations from a past timestep up to the present, and we compute the loss in~\eqref{eq:dynamics_loss} from the discrepancies between these predictions and the actual observations.
We update only $\hat{\mathbf d}$ while keeping the model parameters $\theta^*$ and the weights of $\pi^*$ fixed.
In gradient descent form, the update is
\begin{equation}
\label{eq:online_adaptation}
\hat{\mathbf d}
\leftarrow \hat{\mathbf d}
-\eta_d\nabla_{\hat{\mathbf d}}
\mathcal L_{\mathrm{dyn}}(\theta^*,\hat{\mathbf d}).
\end{equation}
Here, $\eta_d$ is the step size for updating $\hat{\mathbf d}$.
We use the updated estimate for the next policy inference and repeat action generation and online adaptation.

\subsection{Retraining with an Augmented Dataset}
If the dynamics model does not adequately capture interactions encountered during execution, prediction-error-based updates may fail to estimate the physical representation corresponding to the target object.
To improve online estimation, we can retrain both the dynamics model and the VAE that provides its visual representation using action-observation data collected during rollouts, without retraining the policy.
Training the dynamics model on a broader data distribution is expected to improve the accuracy of online adaptation.

First, we add rollout data to the demonstrations collected in step~\ding{172} of Section~\ref{sec:demo_collection}, regardless of task success, to form an augmented dataset.

Next, we retrain the VAE from scratch on images from the augmented dataset, freeze the resulting encoder, and retrain the dynamics model.
During dynamics model retraining, we keep $\{\mathbf d_i^*\}$ fixed and optimize only $\theta$.
Keeping the weights of $\pi^*$ and the learned representations $\{\mathbf d_i^*\}$ fixed allows us to use additional, autonomously collected interaction data without retraining the policy.
Although the rollouts are autonomous, their collection still requires operational effort, such as resetting the environment.

\section{Experimental Setup}
\label{sec:experimental_setup}

\subsection{Manipulation Tasks}
\begin{figure*}[t]
    \centering
    \begin{minipage}[t]{0.32\textwidth}
        \centering
        \includegraphics[width=\linewidth]{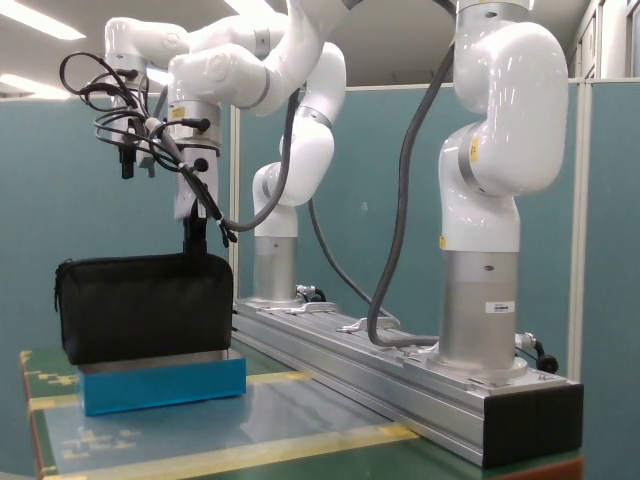}

        \small (a) Pouch Placement
    \end{minipage}
    \hfill
    \begin{minipage}[t]{0.32\textwidth}
        \centering
        \includegraphics[width=\linewidth]{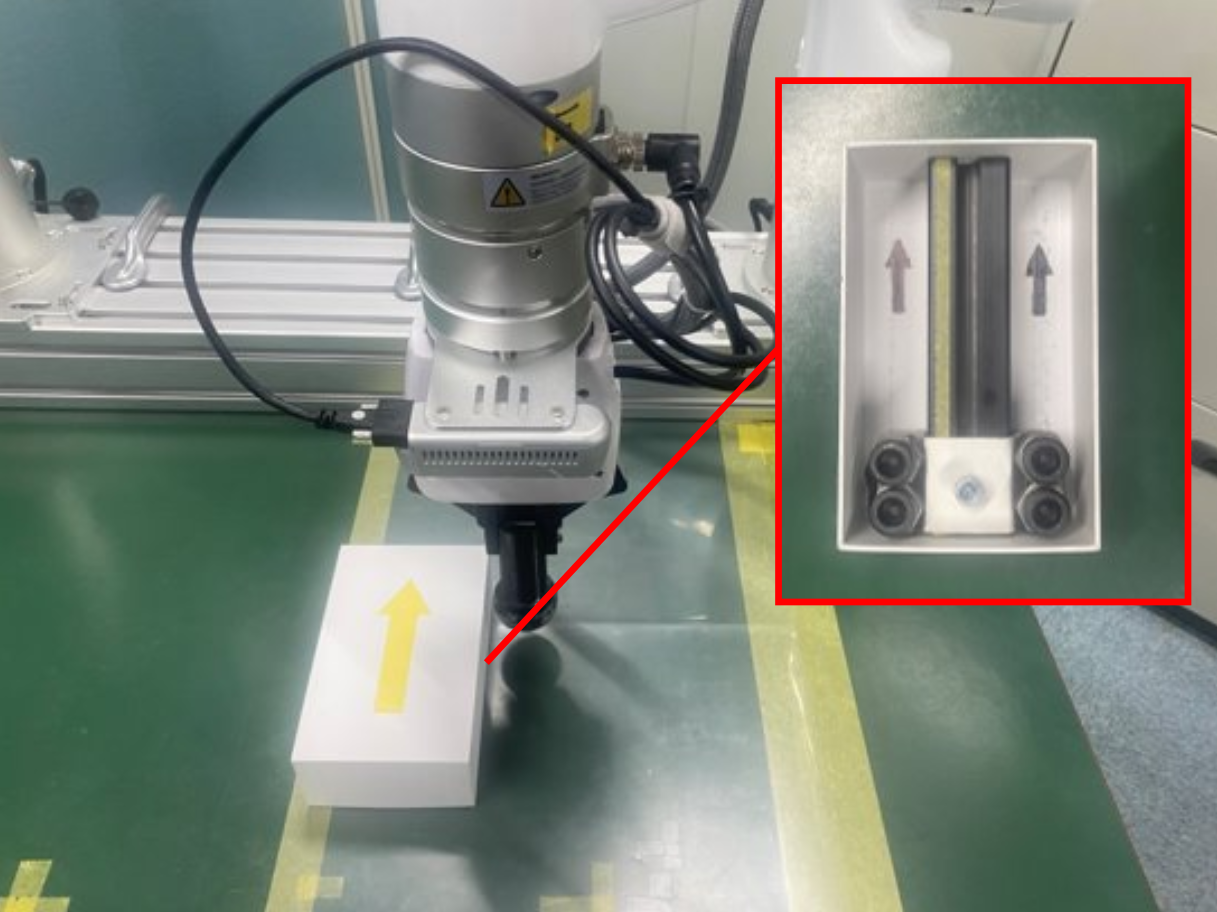}

        \small (b) Box Pushing
    \end{minipage}
    \hfill
    \begin{minipage}[t]{0.32\textwidth}
        \centering
        \includegraphics[width=\linewidth]{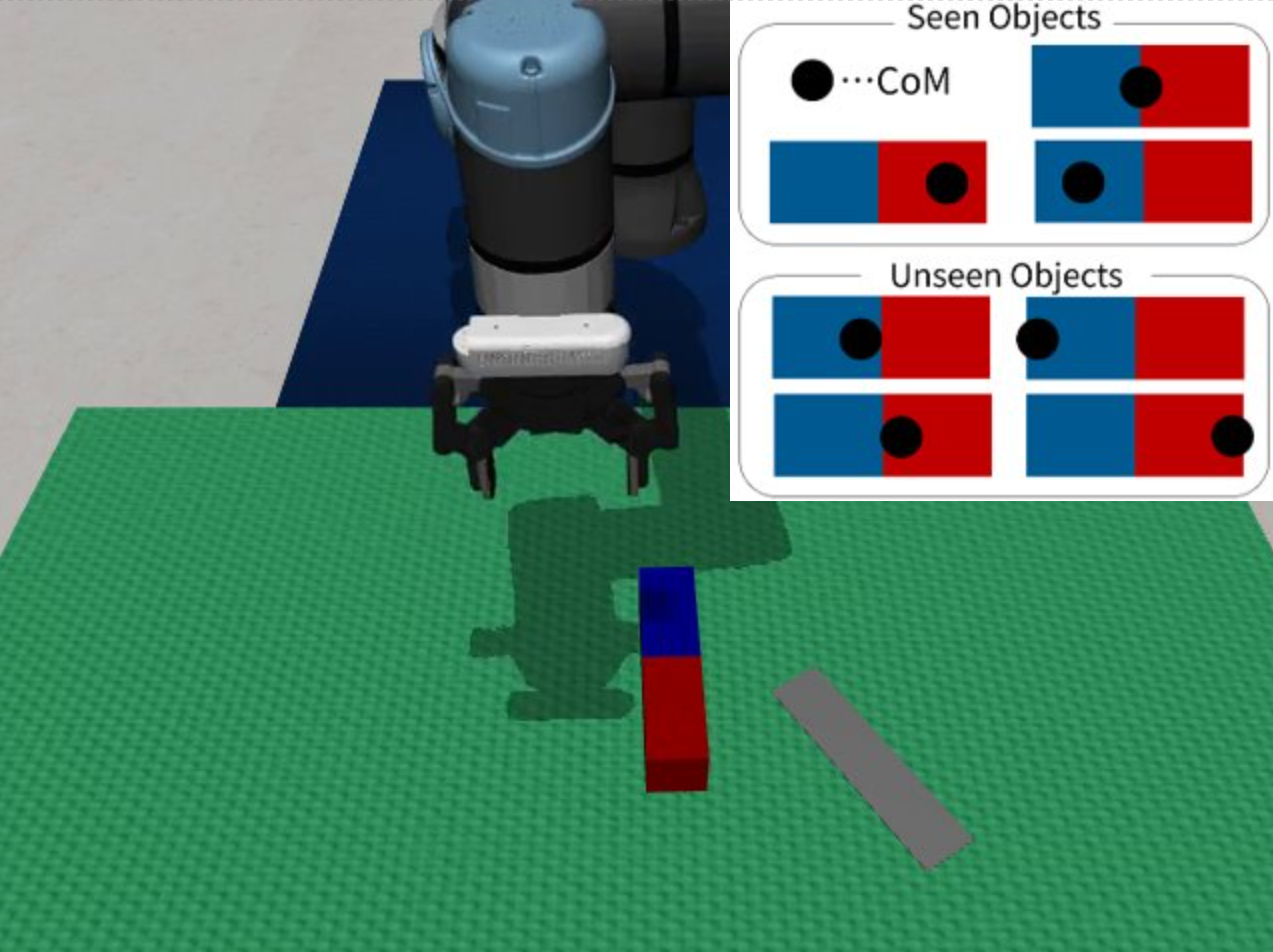}

        \small (c) Horizontal Lifting
    \end{minipage}
    \caption{Manipulation tasks with objects whose centers of mass differ: (a) Pouch Placement and (b) Box Pushing on a real robot, and (c) Horizontal Lifting in simulation.}
    \label{fig:tasks}
\end{figure*}

We evaluate the proposed method on two real-world tasks and use a simulation task to compare dynamics model input representations and observation history lengths (Fig.~\ref{fig:tasks}).
Throughout this paper, unseen objects refer to objects with CoM locations not included in the demonstrations.
We evaluate adaptation to these unseen CoM locations in Box Pushing and Horizontal Lifting.

\noindent (a) Pouch Placement:
The robot must pick up a deformable pouch and place it inside a designated frame within a time limit (Fig.~\ref{fig:tasks}(a)).
We vary the CoM by changing the position of a weight inside the pouch and collect demonstrations for three configurations, with the CoM near the robot, at the center, or far from the robot.
Inserting the pouch into the frame is difficult unless it is level, so the demonstrator gently sets it down and regrasps it only if it tilts.
We evaluate the policies on the same three CoM configurations used for demonstration collection.

\noindent (b) Box Pushing:
The robot must push a rectangular box to the left until its two left corners lie on the yellow tape (Fig.~\ref{fig:tasks}(b)).
In the demonstrations, the box is initially placed at a random orientation between two yellow lines; evaluation uses two fixed initial poses.
In the demonstrations, the operator first aligns the box with the tape, then pushes at a location aligned with its CoM to translate it without rotation.
Task performance therefore depends directly on how quickly the CoM can be inferred during execution.
A slider provides 11 equally spaced weight positions; we collect demonstrations at five positions and evaluate at all 11, including the six unseen positions.

\noindent (c) Horizontal Lifting:
The robot must lift a bar and hold it level above a specified height at the end of a 17-s episode (Fig.~\ref{fig:tasks}(c)).
The bar has red and blue halves of equal length.
We collect demonstrations for three CoM locations: the center of the red half, the center of the bar, and the center of the blue half.
In every demonstration, a teleoperator who does not know the object's physical properties first grasps the bar at its geometric center.
If the bar remains level, the operator maintains that posture; if it tilts, the operator gently sets it down, regrasps near the inferred CoM, and lifts it again.
Thus, the demonstrations explicitly include deliberate adaptation, rather than recovery after an accidental drop.
Evaluation covers these three seen configurations and four additional, unseen CoM locations.

\subsection{Experimental Protocol}
All experiments are conducted using RoboManipBaselines~\cite{Murooka:RoboManipBaselines:IEEEAccess2026}.
The real-world experiments use a UFACTORY xArm7 and one fixed external RealSense D435 camera positioned to view the workspace for each task.
The simulation uses a Universal Robots UR5e manipulator with a parallel gripper.
The control period is approximately 0.129~s for the real robot and 0.096~s for Horizontal Lifting; online updates of the physical representation use the same period.

Table~\ref{tab:dataset_evaluation} summarizes the demonstration and evaluation counts.
Within each task, we use the same initial-pose distribution for all object configurations to avoid bias associated with object type.
We train each method with three random seeds and aggregate success rates across evaluation rollouts.

\begin{table}[t]
    \centering
    \caption{Demonstration and evaluation counts. Each entry gives the count per object configuration, with the total in parentheses. Evaluation counts are per training seed.}
    \label{tab:dataset_evaluation}
    \small
    \begin{tabular}{lcc}
        \toprule
        Task & Demonstrations & Evaluation rollouts \\
        \midrule
        Pouch Placement & 12 (36) & 3 (9) \\
        Box Pushing & 10 (50) & 2 (22) \\
        Horizontal Lifting & 25 (75) & 10 (70) \\
        \bottomrule
    \end{tabular}
\end{table}

\subsection{Compared Methods and Policy Inputs}
Our baseline is a standard Diffusion Policy (DP) trained on pooled demonstrations without distinguishing object types.
Both the baseline and our method use an observation history of two timesteps (approximately 0.2~s).
For the two real-world tasks, we compare this baseline with our method using an image-based dynamics model.
For Horizontal Lifting, we compare four methods: standard DP, DP with a 16-step observation history (approximately 1.5~s), denoted DP (long history), and our method with either a state-based or an image-based dynamics model.
The state-based dynamics model replaces the image features $\mathbf z_t$ with the object's position and orientation from the simulator in both the input history and the prediction targets.
The image-based model uses VAE features as described in Section~\ref{sec:proposed_method}.
These variants differ in the observations used as dynamics model inputs and prediction targets; their policy inputs are identical.

For both real-world tasks, the policies receive RGB images and end-effector position and orientation.
Additional inputs are gripper joint angles for Pouch Placement and measured forces and torques for Box Pushing.
The policies output commands specifying target end-effector position and orientation, with an additional gripper joint-angle command for Pouch Placement.

For Horizontal Lifting, all four methods use a state-based DP conditioned on end-effector position and orientation, gripper joint angles, and object position and orientation from the simulator.
The actions specify target end-effector position and orientation and a gripper joint-angle command.
These observation and action definitions are shared by all methods within each task.

\subsection{Training and Online Adaptation}
Across all tasks, we restrict changes in weight position and the resulting changes in CoM and CoF locations to one dimension.
Accordingly, we use a scalar implicit dynamics parameter.
We set the dynamics model's prediction horizon to $H=14$ steps, corresponding to approximately 1.8~s in the real-world experiments and 1.3~s in simulation.

The image-based dynamics model uses nine-dimensional VAE features, with measured and commanded joint positions, including the gripper, as proprioceptive inputs and action inputs, respectively.
A given end-effector pose can correspond to multiple joint configurations, which appear differently in the fixed external camera (Fig.~\ref{fig:proposed_pipeline}).
Using end-effector pose as input can make image-feature prediction unstable; joint positions avoid this ambiguity.
For Box Pushing, we normalize image features and force/torque measurements separately and weight their prediction errors equally ($\lambda_w=1$ in~\eqref{eq:dynamics_loss}).
% TODO: Confirm the adopted image-based Horizontal Lifting configuration (VAE dimension and model inputs/outputs), and the force/torque prediction loss settings for both variants.

We update the physical representation online using Adam~\cite{kingma2017adammethodstochasticoptimization}, with a step size of $\eta_d=6\times10^{-3}$ for the state-based dynamics model and $\eta_d=8\times10^{-3}$ for the image-based model.

\section{Experimental Results}
\label{sec:experimental_results}

\subsection{Physical Representation Learning and Online Estimation}
\begin{figure*}[!t]
  \centering
  \includegraphics[width=0.87\textwidth]{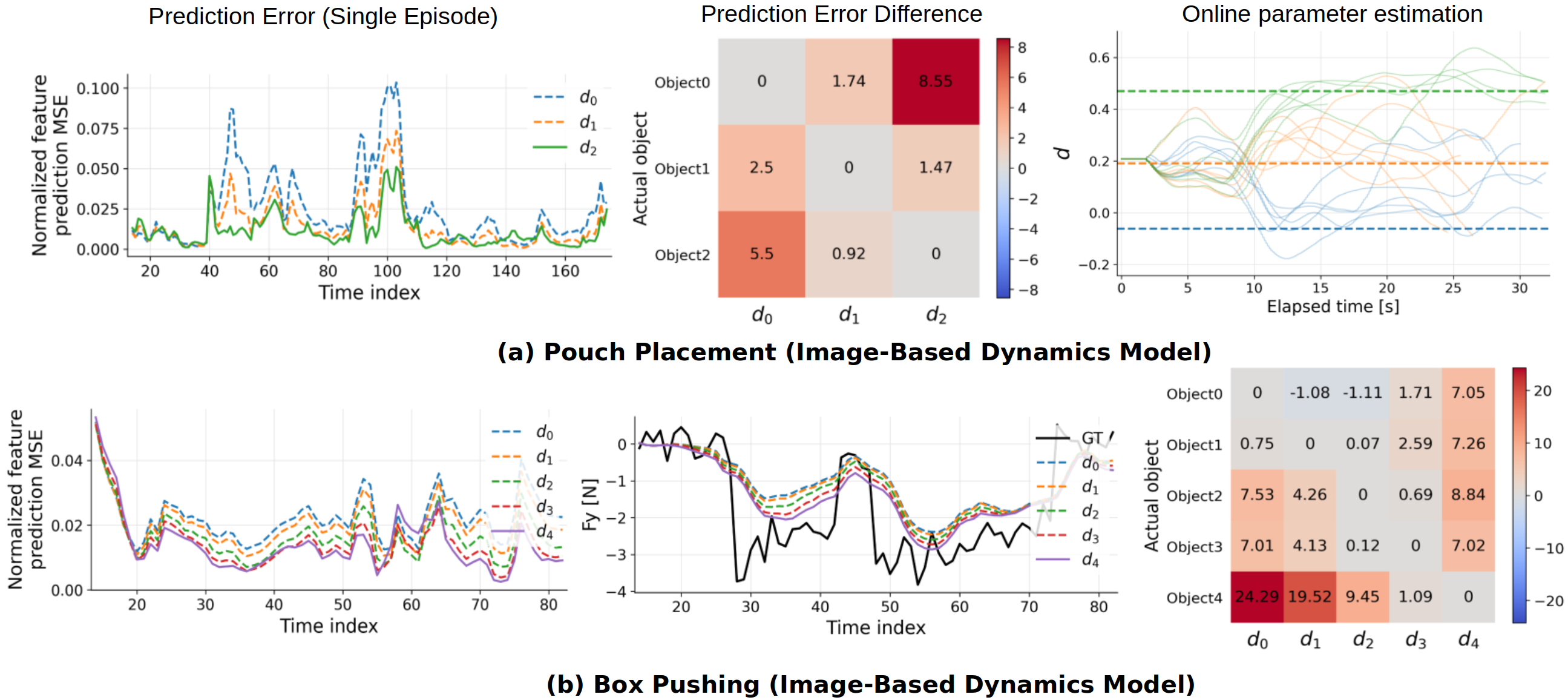}
  \par\smallskip
  \includegraphics[width=0.85\textwidth]{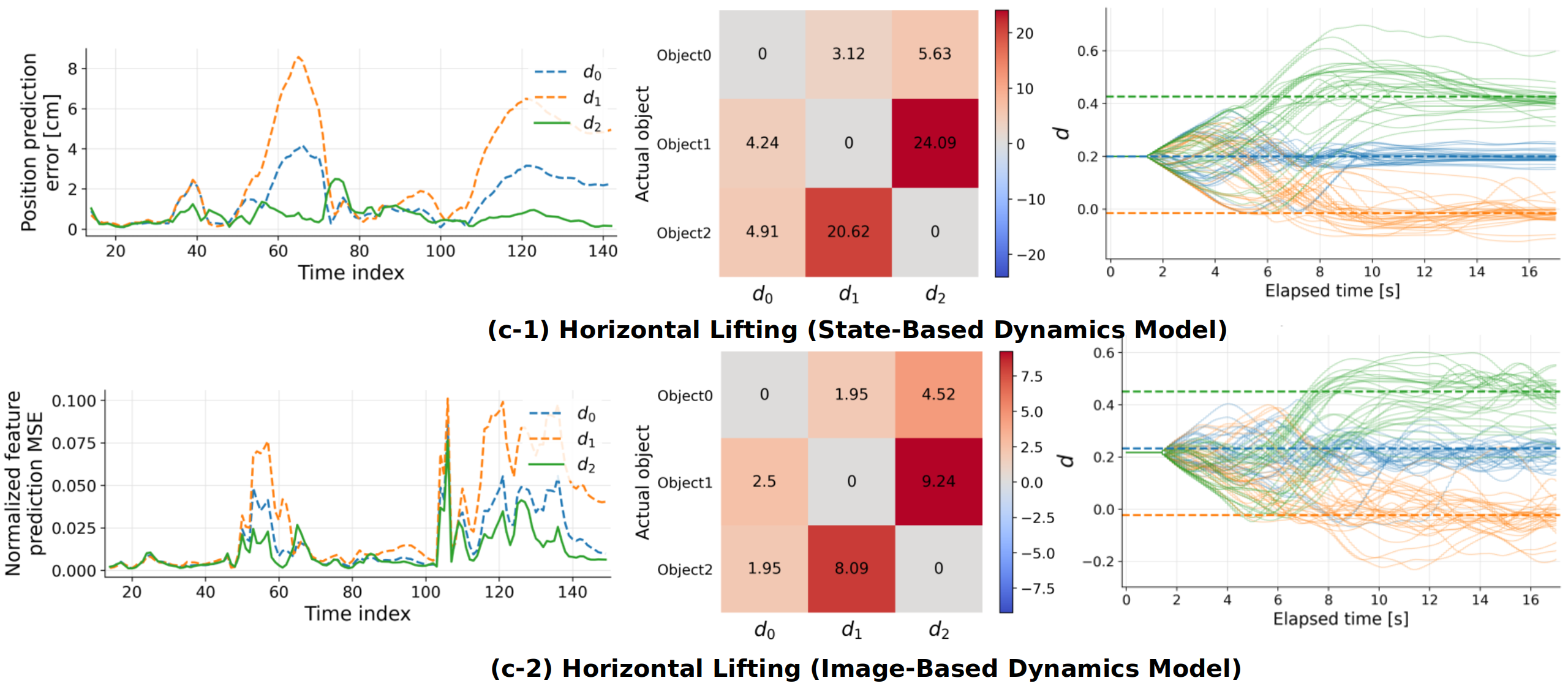}
  \caption{Dynamics model evaluation.
  Left: prediction errors on validation demonstrations excluded from all training. Solid lines use the representation of the actual object; dashed lines use those of other objects. Errors are measured in image features, except in (c-1), which shows position error. Panel (b) also compares predicted and measured force $F_y$ along the pushing direction. Panels (c-1) and (c-2) use the same episode.
  Center: prediction-error differences averaged over all validation data for each object, relative to using its corresponding $\mathbf d_i^*$. Rows indicate the actual objects and columns the input representations. Positive values indicate larger errors when using another object's representation. Values in the heatmaps are reported in units of $10^{-3}$.
  Right: online estimates $\hat{\mathbf d}$ for all evaluation rollouts on seen objects only, with one curve per rollout. Horizontal dashed lines mark the learned, fixed representations $\mathbf d_i^*$; curves and reference lines for the same object share a color.
  For unseen objects with weight positions between those of two seen objects, convergence to values between the corresponding learned representations is desirable.
  Box Pushing trajectories are shown in Fig.~\ref{fig:before_after_retraining}.
  The objects with centered weights have indices 1, 2, and 0 in Pouch Placement, Box Pushing, and Horizontal Lifting, respectively.}
  \label{fig:analyze_dynamics}
\end{figure*}

Fig.~\ref{fig:analyze_dynamics} evaluates whether the learned physical representations $\mathbf d_i^*$ capture differences in object responses and whether $\hat{\mathbf d}$ can be estimated online during rollouts.

The examples in the left column of Fig.~\ref{fig:analyze_dynamics} illustrate lower prediction errors when the model is conditioned on the corresponding $\mathbf d_i^*$, although each example represents only a single episode.

The heatmaps in the center column show that, for Pouch Placement and Horizontal Lifting, the mean prediction error is lower with the representation corresponding to the actual object than with those of other objects.
This suggests that the learned $\mathbf d_i^*$ capture differences in object responses and that the model uses these differences for prediction.
For Box Pushing, however, the corresponding $\mathbf d_i^*$ does not yield the lowest error for some objects, indicating that the representations do not always adequately capture differences between objects.

\subsection{Real-World Experiments}
\begin{figure}[t]
  \centering
  \includegraphics[width=\columnwidth]{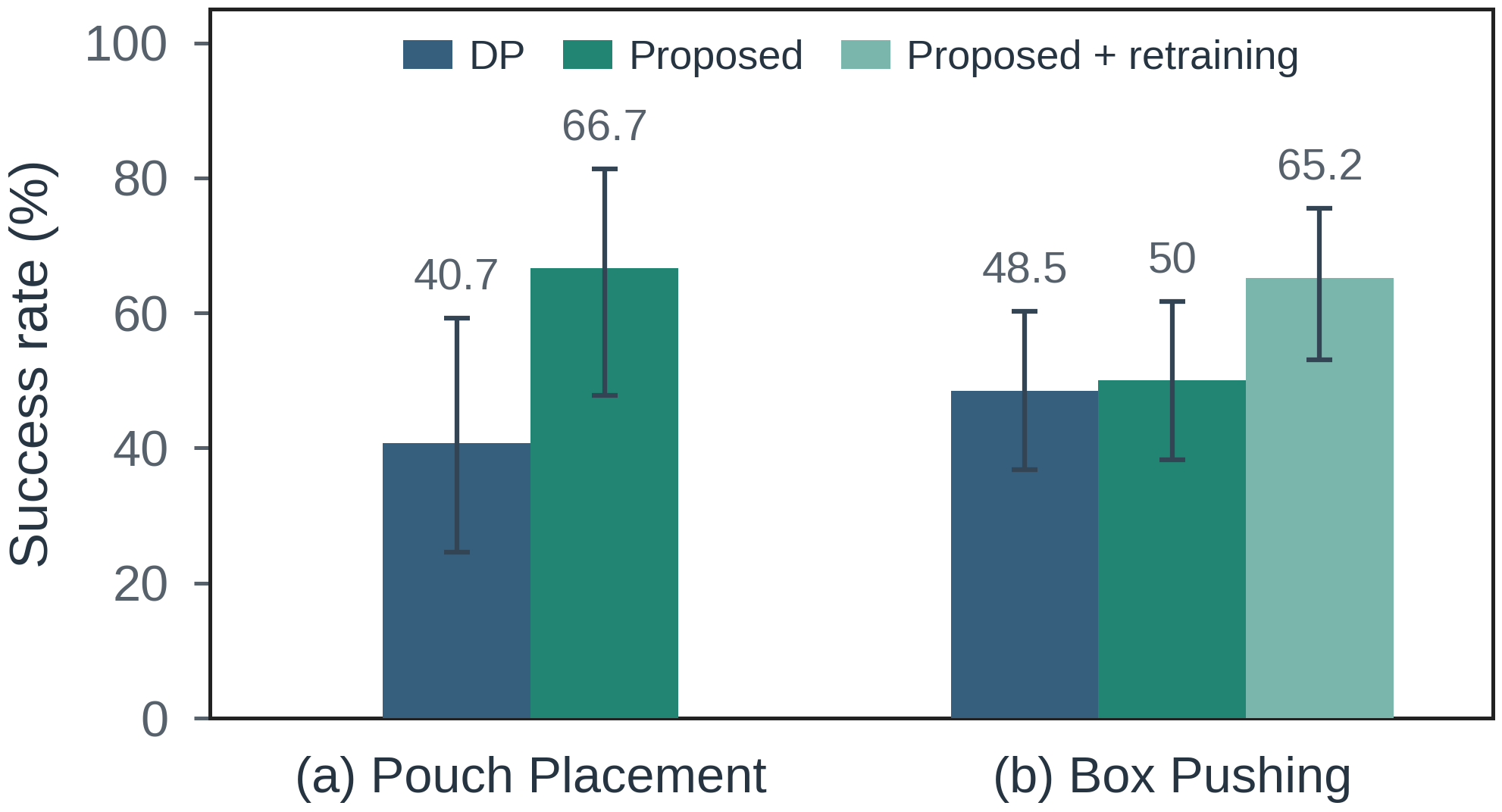}
  \caption{Task success rates for Pouch Placement and Box Pushing on a real robot. ``Proposed + retraining'' denotes retraining the VAE and dynamics model on the augmented dataset while keeping the policy fixed. Error bars denote approximate 95\% Wilson confidence intervals computed from rollouts pooled across three training seeds, treating rollouts as independent. All seeds were evaluated on the same fixed test configurations. Successes per seed: Pouch Placement (of 9), DP 4, 4, 3; Proposed 6, 6, 6. Box Pushing (of 22), DP 9, 11, 12; Proposed 10, 10, 13; Proposed + retraining 13, 13, 17.}
  \label{fig:real_robot_success_rate}
\end{figure}

Fig.~\ref{fig:real_robot_success_rate} summarizes the real-world results.

For Pouch Placement, the proposed method achieves a success rate of 66.7\%, compared with 40.7\% for standard DP, an improvement of approximately 26 percentage points.

In failed DP rollouts, the policy sometimes regrasped the pouch at a location that did not account for its CoM, even after lifting it and setting it down once.

Failures of the proposed method exhibited a different pattern: when the pouch tilted after lifting, $\hat{\mathbf d}$ sometimes updated too slowly and remained near its initial value corresponding to a centered CoM.
The policy then attempted to insert the tilted pouch into the frame without first setting it down to change the grasp location, resulting in failure.
Such attempts to insert a tilted pouch were not observed in standard DP rollouts.
This suggests that, when visual observations and the estimated physical representation favor different actions, the policy may prioritize the behavior associated with $\hat{\mathbf d}$.

For Box Pushing, standard DP and the proposed method trained only on demonstrations achieved similar success rates.
DP's successes were unevenly distributed across objects, favoring those with lower indices, and were more common when little change in pushing location was needed after the initial contact.
Before retraining, the proposed method showed less variation in success across objects than DP.
However, it sometimes changed the contact location unnecessarily even when the initial push was already at an appropriate location.
This behavior was observed when $\hat{\mathbf d}$ took an incorrect value.

We next augmented the original demonstrations with autonomous rollouts of the proposed method on the five training objects, regardless of task success.
Keeping the policy weights and the learned physical representations $\{\mathbf d_i^*\}$ fixed, we reinitialized the VAE and dynamics model and retrained both from scratch on the augmented dataset.

\begin{figure}[t]
  \centering
  \includegraphics[width=\columnwidth]{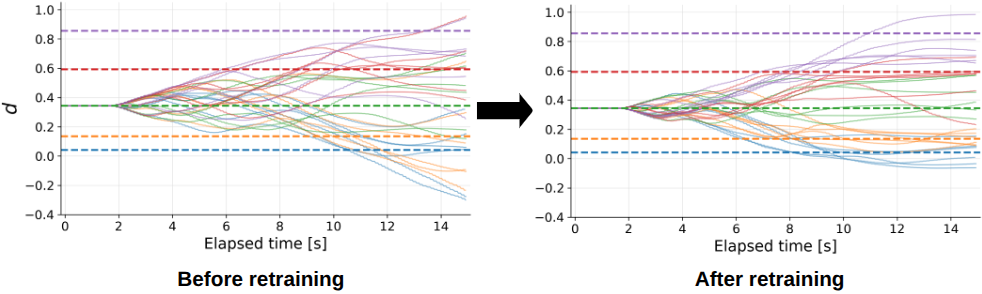}
  \caption{Online estimates $\hat{\mathbf d}$ for Box Pushing before (left) and after (right) reinitializing and retraining the VAE and dynamics model on the augmented dataset. Colors indicate object types. Dashed horizontal lines denote the learned $\mathbf d_i^*$ for objects 0--4 (blue, orange, green, red, and purple, respectively).}
  \label{fig:before_after_retraining}
\end{figure}

After retraining, the success rate increased by approximately 15 percentage points.
Pooled across seeds, success rates on seen and unseen CoM locations were similar for each method: 50.0\% and 47.2\% for DP, 53.3\% and 47.2\% for the proposed method, and 66.7\% and 63.9\% after retraining.
Fig.~\ref{fig:before_after_retraining} compares the trajectories of $\hat{\mathbf d}$ before and after retraining.

As shown in Fig.~\ref{fig:before_after_retraining}, the estimates for some objects exhibit smaller deviations and less variability after retraining.
The initial-position settings for the box were unchanged between the two evaluations, and the added rollouts were collected under the same initial-position conditions as the evaluation.
Improved prediction and estimation under these conditions are therefore expected from the additional data.
This comparison examines how improved physical-property estimation affects behavior and task success while the policy remains fixed.
Because retraining uses observation--action sequences regardless of task outcome, failed rollouts can also contribute to model learning.

After retraining, rollouts exhibited fewer unnecessary changes in contact location and more successful episodes.
These results suggest that more stable physical-property estimation reduced unnecessary contact changes and contributed to the increase in success rate.

Across both real-world tasks, episodes in which $\hat{\mathbf d}$ took an incorrect value tended to fail.
This suggests that $\hat{\mathbf d}$ influences action selection as a policy conditioning input and that accurate estimation is important for generating appropriate actions.
The results therefore suggest that task success depends on the predictive accuracy of the dynamics model used for online physical-property estimation.

\subsection{Simulation Experiments}
Table~\ref{tab:simulation_success_rate} reports the results for Horizontal Lifting.
For each training seed, we evaluate 30 rollouts on seen objects and 40 on unseen objects; the total success rate is computed over all 70 rollouts.

The proposed method achieves a higher mean overall success rate than both standard DP and DP (long history), with either the state-based or the image-based dynamics model.
Compared with standard DP, DP (long history) achieves the same success rate on seen objects but a lower rate on unseen objects, indicating that simply extending the observation history does not improve performance in this experiment.

As shown in Fig.~\ref{fig:analyze_dynamics}(c-1) and (c-2), both dynamics models achieve the lowest mean prediction error when conditioned on the representation $\mathbf d_i^*$ corresponding to the actual object.

The online estimates tend to vary more with the image-based dynamics model than with the state-based model, yet the two variants achieve similar overall success rates.

On unseen objects, both variants outperform the baselines in mean success rate, but their rates remain around 30\%.
Adapting to CoM locations absent from the demonstrations therefore remains challenging.

\begin{table}[t]
  \centering
  \caption{Success rates (\%) for Horizontal Lifting, reported as mean $\pm$ sample standard deviation across three training seeds.}
  \label{tab:simulation_success_rate}
  \small
  \setlength{\tabcolsep}{3pt}
  \renewcommand{\arraystretch}{1.15}
  \begin{tabular}{@{}lccc@{}}
    \toprule
    Method & \shortstack{Seen\\Objects} & \shortstack{Unseen\\Objects} & Total \\
    \midrule
    DP & $85.6 \pm 5.1$ & $21.7 \pm 5.2$ & $49.0 \pm 0.8$ \\
    DP (long history) & $85.6 \pm 5.1$ & $6.7 \pm 2.9$ & $40.5 \pm 3.0$ \\
    \addlinespace[2pt]
    Proposed (state-based) & $\mathbf{96.7 \pm 3.3}$ & $30.0 \pm 6.6$ & $\mathbf{58.6 \pm 2.9}$ \\
    Proposed (image-based) & $92.2 \pm 3.8$ & $\mathbf{32.5 \pm 2.5}$ & $58.1 \pm 0.8$ \\
    \bottomrule
  \end{tabular}
\end{table}

\section{Limitations}
Our method does not always reliably estimate the physical representation $\hat{\mathbf d}$ from brief interactions.
In the real-world experiments, delayed or inaccurate estimates were associated with inappropriate insertion attempts and unnecessary changes in contact location.
In simulation, success rates on CoM locations absent from the demonstrations remained around 30\%, indicating limited adaptation to unseen physical properties.
Effective policy conditioning therefore requires rapid and accurate estimation of $\hat{\mathbf d}$ from observations during task execution.
Future work should investigate whether pretrained visual representations and dynamics models can improve prediction accuracy and the stability of online estimation, and whether these improvements lead to better adaptation to unseen CoM locations.

Our experiments primarily varied weight positions in one dimension and used a scalar physical representation $\mathbf d_i$.
We have not evaluated conditions in which the CoM varies in three dimensions or tasks that require switching behaviors according to object mass.
Further evaluation with multidimensional physical representations is needed to determine whether the method can learn adaptive behaviors for these variations in physical properties.

\section{Conclusion}
We presented an imitation learning method that learns latent representations of physical properties from demonstrations to predict object responses and updates these representations online using prediction errors during task execution.
The same demonstrations provide supervision for predicting object responses and learning adaptive actions, including regrasping and changes in contact location.
Conditioning the fixed Diffusion Policy on the online estimate allows information acquired through interaction to guide these behaviors.
Real-world and simulation experiments showed that the learned representations capture differences in object responses and that conditioning on their online estimates strongly influences action selection and affects adaptation performance.
Retraining the VAE and dynamics model on demonstrations augmented with autonomous rollout data further improved task success while keeping the policy fixed.
Our evaluation was limited to one-dimensional variation in mass distribution, with Pouch Placement evaluated only on seen configurations, and adaptation to unseen configurations was not strong in either the real-world or the simulation experiments.
Nevertheless, these results suggest that online adaptation through latent physical representations can improve imitation-based manipulation of objects whose physical properties cannot be distinguished visually.


%%%%%%%%%%%%%%%%%%%%%%%%%%%%%%%%%%%%%%%%%%%%%%%%%%%%%%%%%%%%%%%%%%%%%%%%%%%%%%%%

%%%%%%%%%%%%%%%%%%%%%%%%%%%%%%%%%%%%%%%%%%%%%%%%%%%%%%%%%%%%%%%%%%%%%%%%%%%%%%%%

%%%%%%%%%%%%%%%%%%%%%%%%%%%%%%%%%%%%%%%%%%%%%%%%%%%%%%%%%%%%%%%%%%%%%%%%%%%%%%%%

%%%%%%%%%%%%%%%%%%%%%%%%%%%%%%%%%%%%%%%%%%%%%%%%%%%%%%%%%%%%%%%%%%%%%%%%%%%%%%%%

\bibliographystyle{IEEEtran}
\bibliography{references,quick_reference,IEEEabrv,IEEEexample}

@inproceedings{Longhini:EDO-Net:ICRA2023,
  title = {{EDO-Net}: Learning Elastic Properties of Deformable Objects from Graph Dynamics},
  doi = {10.1109/icra48891.2023.10161234},
  booktitle = {2023 IEEE International Conference on Robotics and Automation (ICRA)},
  author = {Longhini, Alberta and Moletta, Marco and Reichlin, Alfredo and Welle, Michael C. and Held, David and Erickson, Zackory and Kragic, Danica},
  year = {2023},
  pages = {3875--3881},
}

@inproceedings{Evans:IIDA:ICRA2022,
  title = {Context is Everything: Implicit Identification for Dynamics Adaptation},
  doi = {10.1109/icra46639.2022.9812119},
  booktitle = {2022 International Conference on Robotics and Automation (ICRA)},
  author = {Evans, Ben and Thankaraj, Abitha and Pinto, Lerrel},
  year = {2022},
  pages = {2642--2648},
}

@InProceedings{Murooka:EXI-Net:PMLR2021,
  title = 	 {{EXI-Net}: EXplicitly/Implicitly Conditioned Network for Multiple Environment Sim-to-Real Transfer},
  author =       {Murooka, Takayuki and Hamaya, Masashi and Drigalski, Felix von and Tanaka, Kazutoshi and Ijiri, Yoshihisa},
  booktitle = 	 {Proceedings of the 2020 Conference on Robot Learning},
  pages = 	 {1221--1230},
  year = 	 {2021},
  volume = 	 {155},
  publisher =    {PMLR}
}

@inproceedings{Aoyama:FewShotHaptic:ICRA2024,
  title = {Few-Shot Learning of Force-Based Motions From Demonstration Through Pre-training of Haptic Representation},
  doi = {10.1109/icra57147.2024.10610502},
  booktitle = {2024 IEEE International Conference on Robotics and Automation (ICRA)},
  author = {Aoyama, Marina Y. and Moura, João and Saito, Namiko and Vijayakumar, Sethu},
  year = {2024},
  pages = {12839--12845},
}

@article{Saito:SelectAndUseTools:RAL2021,
  title = {How to Select and Use Tools? : Active Perception of Target Objects Using Multimodal Deep Learning},
  volume = {6},
  doi = {10.1109/lra.2021.3062004},
  number = {2},
  journal = {IEEE Robotics and Automation Letters},
  author = {Saito, Namiko and Ogata, Tetsuya and Funabashi, Satoshi and Mori, Hiroki and Sugano, Shigeki},
  year = {2021},
  pages = {2517--2524},
}

@inproceedings{SobolMark:BPP:RSS2026,
  author = {Max Sobol Mark AND Maria Attarian AND Chuyuan Fu AND Debidatta Dwibedi AND Jacky Liang AND Dhruv Shah AND Aviral Kumar},
  title = {{Long-Context Robot Imitation Learning by Focusing on Key History Frames}},
  booktitle = {Proceedings of Robotics: Science and Systems (RSS)},
  year = {2026},
  address = {Sydney, Australia},
  doi = {10.15607/RSS.2026.XXII.201},
}

@article{An:FeedbackWorldModel:arXiv2026,
  title = {Feedback World Model Enables Precise Guidance of Diffusion Policy},
  author = {Tuo An and Jindou Jia and Gen Li and Jingliang Li and Chuhao Zhou and Pengfei Liu and Bofan Lyu and Jiaqi Bai and Xinying Guo and Geng Li and Jianfei Yang},
  year = {2026},
  journal = {arXiv preprint arXiv:2605.15705},
}

@inproceedings{Lyu:DyWA:ICCV2025,
  author = {Lyu, Jiangran and Li, Ziming and Shi, Xuesong and Xu, Chaoyi and Wang, Yizhou and Wang, He},
  title = {{DyWA}: Dynamics-adaptive World Action Model for Generalizable Non-prehensile Manipulation},
  booktitle = {Proceedings of the IEEE/CVF International Conference on Computer Vision (ICCV)},
  year = {2025},
  pages = {11058-11068},
}

@article{Yuan:AdaWorldPolicy:arXiv2026,
  title = {{AdaWorldPolicy}: World-Model-Driven Diffusion Policy with Online Adaptive Learning for Robotic Manipulation},
  author = {Ge Yuan and Qiyuan Qiao and Jing Zhang and Dong Xu},
  year = {2026},
  journal = {arXiv preprint arXiv:2602.20057},
}

@article{Ali:Cosmos:arXiv2025,
  title = {World Simulation with Video Foundation Models for Physical {AI}},
  author = {Arslan Ali and Junjie Bai and Maciej Bala and Yogesh Balaji and Aaron Blakeman and Tiffany Cai and others},
  year = {2025},
  journal = {arXiv preprint arXiv:2511.00062},
}

@article{Chi:DiffusionPolicy:IJRR2025,
  title = {Diffusion policy: Visuomotor policy learning via action diffusion},
  volume = {44},
  doi = {10.1177/02783649241273668},
  number = {10-11},
  journal = {The International Journal of Robotics Research},
  author = {Chi, Cheng and Xu, Zhenjia and Feng, Siyuan and Cousineau, Eric and Du, Yilun and Burchfiel, Benjamin and Tedrake, Russ and Song, Shuran},
  year = {2025},
  pages = {1684--1704},
}

@inproceedings{Aoyama:PokeAndStrike:CoRL2025,
  title={Poke and Strike: Learning Task-Informed Exploration Policies},
  author={Marina Y. Aoyama and Joao Moura and Juan Del Aguila Ferrandis and Sethu Vijayakumar},
  booktitle={9th Annual Conference on Robot Learning},
  year={2025},
}

@inproceedings{Hou:AdaptiveCompliancePolicy:ICRA2025,
  title={Adaptive compliance policy: Learning approximate compliance for diffusion guided control},
  author={Hou, Yifan and Liu, Zeyi and Chi, Cheng and Cousineau, Eric and Kuppuswamy, Naveen and Feng, Siyuan and Burchfiel, Benjamin and Song, Shuran},
  booktitle={2025 IEEE International Conference on Robotics and Automation (ICRA)},
  pages={4829--4836},
  year={2025},
  organization={IEEE}
}

@inproceedings{Choi:UMI-FT:ICRA2026,
  title={In-the-Wild Compliant Manipulation with {UMI-FT}},
  author={Hojung Choi and Yifan Hou and Chuer Pan and Seongheon Hong and Austin Patel and Xiaomeng Xu and Mark R. Cutkosky and Shuran Song},
  year={2026},
  booktitle={2026 IEEE International Conference on Robotics and Automation (ICRA)},
  organization={IEEE},
}

@article{Hogan:HybridMPC:IJRR2020,
author = {Francois R Hogan and Alberto Rodriguez},
title ={Reactive planar non-prehensile manipulation with hybrid model predictive control},

journal = {The International Journal of Robotics Research},
volume = {39},
number = {7},
pages = {755-773},
year = {2020},
doi = {10.1177/0278364920913938},    
}

@ARTICLE{Murooka:RoboManipBaselines:IEEEAccess2026,
  author={Murooka, Masaki and Motoda, Tomohiro and Nakajo, Ryoichi and Oh, Hanbit and Makihara, Koshi and Shirai, Keisuke and Ogata, Tetsuya and Domae, Yukiyasu},
  journal={IEEE Access}, 
  title={{RoboManipBaselines}: A Unified Framework for Imitation Learning in Robotic Manipulation Across Real and Simulation Environments},
  year={2026},
  volume={14},
  number={},
  pages={97896-97909},
}

@INPROCEEDINGS{Murooka:FeasibilityEvaluation:ICRA2017,
  author={Murooka, Masaki and Nozawa, Shunichi and Kakiuchi, Yohei and Okada, Kei and Inaba, Masayuki},
  booktitle={2017 IEEE International Conference on Robotics and Automation (ICRA)}, 
  title={Feasibility evaluation of object manipulation by a humanoid robot based on recursive estimation of the object's physical properties}, 
  year={2017},
  volume={},
  number={},
  pages={4082-4089}
}

@ARTICLE{Ichiwara:SpatialAttentionPointNetwork:arXiv2021,
  author = {Ichiwara, Hideyuki and Ito, Hiroshi and Yamamoto, Kenjiro and Mori, Hiroki and Ogata, Tetsuya},
  title = {Spatial Attention Point Network for Deep-learning-based Robust Autonomous Robot Motion Generation},
  journal = {arXiv preprint arXiv:2103.01598},
  year = {2021}
}

@ARTICLE{Zhou:MTIL:RAL2025,
  author = {Zhou, Yulin and Lin, Yuankai and Peng, Fanzhe and Chen, Jiahui and Huang, Kaiji and Yang, Hua and Yin, Zhouping},
  title = {{MTIL}: Encoding Full History with Mamba for Temporal Imitation Learning},
  journal = {IEEE Robotics and Automation Letters},
  year = {2025},
  volume = {10},
  number = {11},
  pages = {11761--11767},
  doi = {10.1109/LRA.2025.3615520}
}

@inproceedings{kingma2022autoencodingvariationalbayes,
  title = {Auto-Encoding Variational Bayes},
  author = {Diederik P Kingma and Max Welling},
  booktitle = {International Conference on Learning Representations (ICLR)},
  year = {2014},
}

@inproceedings{kingma2017adammethodstochasticoptimization,
  title = {Adam: A Method for Stochastic Optimization},
  author = {Diederik P. Kingma and Jimmy Ba},
  booktitle = {International Conference on Learning Representations (ICLR)},
  year = {2015},
}

\end{document}